\pdfoutput=1

\documentclass[11pt]{article}

\usepackage[]{acl}

\usepackage{times}
\usepackage{latexsym}

\usepackage[T1]{fontenc}

\usepackage[utf8]{inputenc}

\usepackage{microtype}

\usepackage{amsmath}

\DeclareMathOperator*{\argmin}{arg\,min}
\usepackage{amssymb}
\usepackage{subcaption}
\usepackage{caption}
\usepackage{multicol}
\usepackage{multirow}
\usepackage{booktabs}
\usepackage{graphicx}
\usepackage{float}
\usepackage{efbox}
\usepackage{pifont}
\usepackage{xspace,mfirstuc,tabulary}
\usepackage{array}
\usepackage{arydshln}
\def\@fnsymbol#1{\ensuremath{\ifcase#1\or *\or \dagger\or \ddagger\or
   \mathsection\or \mathparagraph\or \|\or **\or \dagger\dagger
   \or \ddagger\ddagger \else\@ctrerr\fi}}
\newcommand{\ssymbol}[1]{^{\@fnsymbol{#1}}}

\title{Fine-tuning Pre-trained Language Models for Few-shot Intent Detection: Supervised Pre-training and Isotropization}

\author{
Haode Zhang$^1$ \quad Haowen Liang$^1$ \quad Yuwei Zhang$^2$ \\  \bf Liming Zhan$^1$ \quad Xiaolei Lu$^3$ \quad \bf Albert Y.S. Lam$^4$ \quad  Xiao-Ming Wu$^1$\Thanks{~ Corresponding author.} \\ 
Department of Computing, The Hong Kong Polytechnic University, Hong Kong S.A.R.$^1$ \\
University of California, San Diego$^2$ \quad Nanyang Technological University, Singapore$^3$ \\
Fano Labs, Hong Kong S.A.R.$^4$ \\
{\tt \small \{haode.zhang,michaelhw.liang,lmzhan.zhan\}@connect.polyu.hk, zhangyuwei.work@gmail.com} \\
{\tt \small xiao-ming.wu@polyu.edu.hk, xiaolei.lu@ntu.edu.sg, albert@fano.ai} \\
}

\begin{document}
\maketitle
\begin{abstract} 
It is challenging to train a good intent classifier for a task-oriented dialogue system with only a few annotations. Recent studies have shown that fine-tuning pre-trained language models with a small amount of labeled utterances from public benchmarks in a supervised manner is extremely helpful. However, we find that supervised pre-training yields an anisotropic feature space, which may suppress the expressive power of the semantic representations. Inspired by recent research in isotropization, we propose to improve supervised pre-training by regularizing the feature space towards isotropy. We propose two regularizers based on contrastive learning and correlation matrix respectively, and demonstrate their effectiveness through extensive experiments. Our main finding is that it is promising to regularize supervised pre-training with isotropization to further improve the performance of few-shot intent detection. 
The source code can be found at \url{https://github.com/fanolabs/isoIntentBert-main}.
\end{abstract}

\section{Introduction}
Intent detection is a core module of task-oriented dialogue systems. Training a well-performing intent classifier with only a few annotations, i.e., few-shot intent detection, is of great practical value. Recently, this problem has attracted considerable attention~\cite{vulic2021convfit, zhang-etal-2021-shot, dopierre2020few} but remains a challenge.



To tackle few-shot intent detection, earlier works employ induction network~\cite{geng2019few}, generation-based methods~\cite{xia2020composed}, metric learning~\cite{nguyen2020dynamic}, and self-training~\cite{dopierre2020few}, to design sophisticated algorithms. 
Recently, pre-trained language models (PLMs) have emerged as a simple yet promising solution to a wide spectrum of natural language processing (NLP) tasks, triggering the surge of PLM-based solutions for few-shot intent detection~\cite{wu2020tod, zhang-etal-2021-effectiveness-pre, zhang-etal-2021-shot, vulic2021convfit, zhang-etal-2021-shot}, which typically fine-tune PLMs on conversation data. 

A PLM-based fine-tuning method~\cite{zhang-etal-2021-effectiveness-pre}, called IntentBERT, utilizes a small amount of labeled utterances from public intent datasets to fine-tune PLMs with a standard classification task, which is referred to as \emph{supervised pre-training}. Despite its simplicity, supervised pre-training has been shown extremely useful for few-shot intent detection even when the target data and the data used for fine-tuning are very different in semantics. However, as will be shown in Section~\ref{section: pilot experiments, fine-tuning -> anisotropy}, IntentBERT suffers from severe anisotropy, an undesirable property of PLMs~\cite{gao2019representation, ethayarajh2019contextual, li2020sentence}.

\begin{figure*}[t]
\centering
    \centering
    \includegraphics[width=1.0\linewidth]{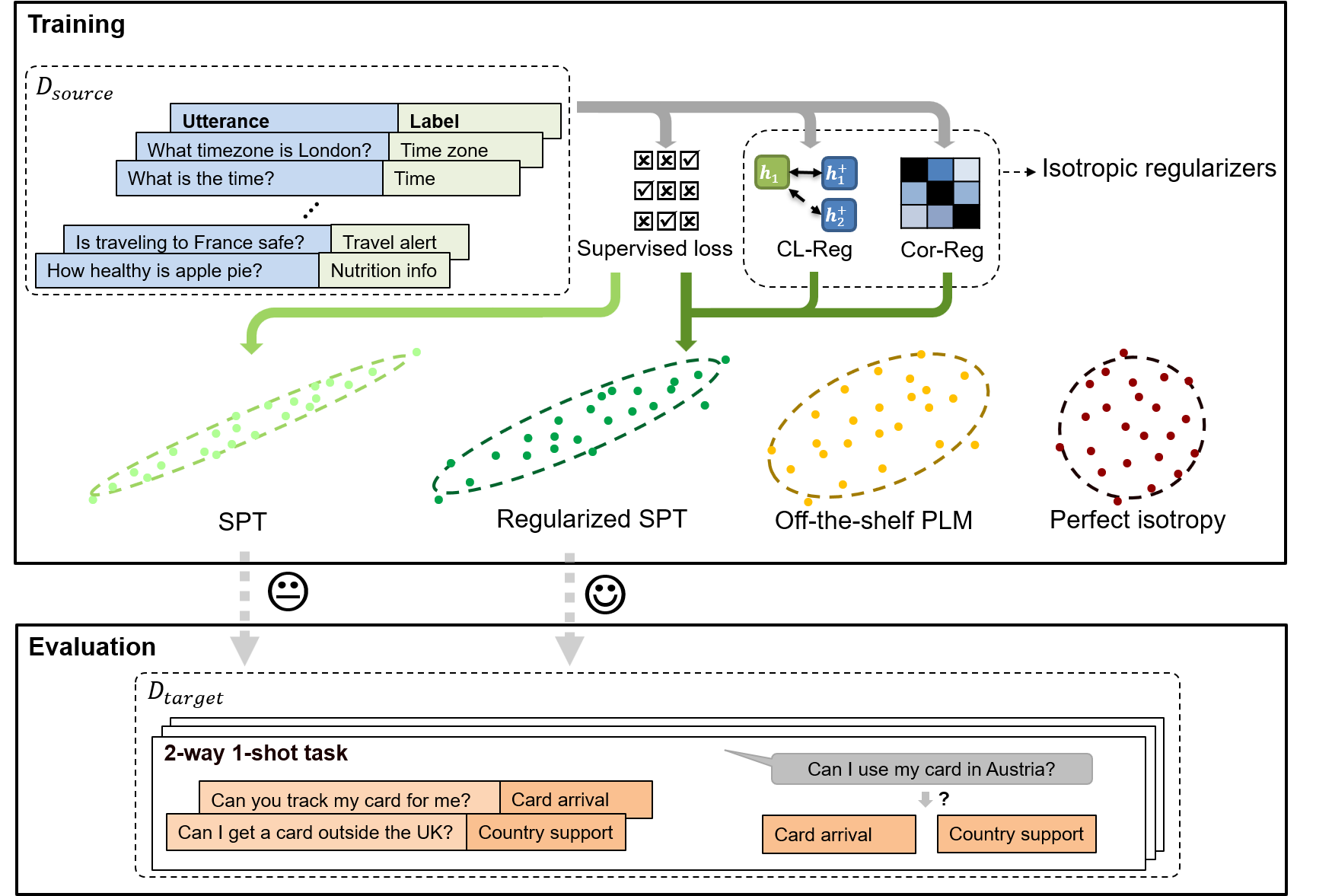}
\caption{Illustration of our proposed regularized supervised pre-training. SPT denotes supervised pre-training (fine-tuning an off-the-shelf PLM on a set of labeled utterances), which makes the feature space more anisotropic. CL-Reg and Cor-Reg are designed to regularize SPT and increase the isotropy of the feature space, which leads to better performance on few-shot intent detection. 
}
\label{figure: idea_illustration}
\end{figure*}

Anisotropy is a geometric property that semantic vectors fall into a narrow cone. It has been identified as 
a crucial factor for the sub-optimal performance of PLMs on a variety of downstream tasks~\cite{gao2019representation, arora-etal-2016-latent, cai2020isotropy, ethayarajh2019contextual}, which is also known as the representation degeneration problem~\cite{gao2019representation}. Fortunately, isotropization techniques can be applied to adjust the embedding space and yield significant performance improvement in many tasks~\cite{su2021whitening, rajaee2021cluster}.

Hence, this paper aims to answer the question:
\begin{itemize}
    \item Can we improve supervised pre-training via \emph{isotropization} for few-shot intent detection?
\end{itemize}
Many isotropization techniques have been developed based on transformation~\cite{su2021whitening, huang-etal-2021-whiteningbert-easy}, contrastive learning~\cite{gao2021simcse}, and top principal components elimination~\cite{DBLP:conf/iclr/MuV18}. However, these methods are designed for off-the-shelf PLMs. When applied on PLMs that have been fine-tuned on some NLP task such as semantic textual similarity or intent classification, they may introduce an adverse effect, as observed in \citet{rajaee2021isotropy} and our pilot experiments.

In this work, we propose to regularize supervised pre-training with isotropic regularizers.
As shown in Fig.~\ref{figure: idea_illustration}, we devise two regularizers, a contrastive-learning-based regularizer (CL-Reg) and a correlation-matrix-based regularizer (Cor-Reg), each of which can increase the isotropy of the feature space  during supervised training. Our empirical study shows that the regularizers can significantly improve the performance of standard supervised training, 
and better performance can often be achieved when they are combined.

The contributions of this work are three-fold:
\begin{itemize}
    \item We present the first study on the isotropy property of PLMs for few-shot intent detection, shedding light on the interaction of supervised pre-training and isotropization. 
    \item We improve supervised pre-training by devising two simple yet effective regularizers to 
    increase the isotropy of the feature space.
    \item We conduct a  comprehensive evaluation and analysis to validate the effectiveness of the proposed approach.
\end{itemize}


\section{Related Works}

\subsection{Few-shot Intent Detection}
With a surge of interest in few-shot learning~\cite{finn2017model, vinyals2016matching, snell2017prototypical}, few-shot intent detection has started to receive attention. Earlier works mainly focus on model design, using capsule network~\cite{geng2019few}, variational autoencoder~\cite{xia2020composed}, or metric functions~\cite{NAACL18, nguyen2020dynamic}. Recently, PLMs-based methods have shown promising performance in a variety of NLP tasks and become the model of choice for few-shot intent detection. 
\citet{zhang-etal-2020-discriminative} cast few-shot intent detection into a natural language inference~(NLI) problem and fine-tune PLMs on NLI datasets. \citet{zhang-etal-2021-shot} propose to fine-tune PLMs on unlabeled utterances by contrastive learning. \citet{zhang-etal-2021-effectiveness-pre} leverage a small set of public annotated intent detection benchmarks to fine-tune PLMs with standard supervised training and observe promising performance on cross-domain few-shot intent detection. Meanwhile, the study of few-shot intent detection has been extended to other settings including semi-supervised learning~\cite{dopierre2020few, dopierre-etal-2021-protaugment}, generalized setting~\cite{nguyen2020dynamic}, multi-label classification~\cite{hou2020few}, and incremental learning~\cite{xia-etal-2021-incremental}. In this work, we consider standard few-shot intent detection, following the setup of \citet{zhang-etal-2021-effectiveness-pre} and aiming to improve supervised pre-training with isotropization.
\subsection{Further  Pre-training PLMs with Dialogue Corpora}
Recent works have shown that further pre-training off-the-shelf PLMs using dialogue corpora~\cite{henderson-etal-2019-training,peng2020few,peng2020soloist} are beneficial for task-oriented downstream tasks such as intent detection.
Specifically, TOD-BERT \cite{wu2020tod} conducts self-supervised learning on diverse task-oriented dialogue corpora. ConvBERT \cite{mehri2020dialoglue} is pre-trained on a $700$ million open-domain dialogue corpus.
\citet{vulic2021convfit} propose a two-stage procedure: adaptive
conversational fine-tuning followed by task-tailored conversational fine-tuning.
In this work, we follow \citet{zhang-etal-2021-effectiveness-pre} to further pre-train PLMs using a small amount of labeled utterances from public intent detection benchmarks. 


\subsection{Anisotropy of PLMs}
Isotropy is a key geometric property of the semantic space of PLMs. Recent studies identify the anisotropy problem of PLMs~\cite{cai2020isotropy,  ethayarajh2019contextual, DBLP:conf/iclr/MuV18, rajaee2021isotropy}, which is also known as the representation degeneration problem~\cite{gao2019representation}: word embeddings occupy a narrow cone, which suppresses the expressiveness of PLMs. To resolve the problem, various methods have been proposed, including spectrum control~\cite{wang2019improving}, flow-based mapping~\cite{li2020sentence}, whitening transformation~\cite{su2021whitening, huang-etal-2021-whiteningbert-easy}, contrastive learning~\cite{gao2021simcse}, and cluster-based methods~\cite{rajaee2021cluster}. 
Despite their effectiveness, these methods are designed for off-the-shelf PLMs. 
The interaction between isotropization and fine-tuning PLMs remains under-explored. A most recent work by \citeauthor{rajaee2021does} shows that there might be a conflict between the two operations for the semantic textual similarity~(STS) task. 
On the other hand, \citet{Zhou2021IsoBNFB} propose to fine-tune PLMs with isotropic batch normalization on some supervised tasks, but it requires a large amount of training data. In this work, we study the interaction between isotropization and supervised pre-training (fine-tuning) PLMs on intent detection tasks. 

\section{Pilot  Study}
\label{section: pilot experiments}
Before introducing our approach, we present pilot experiments to gain some insights into the interaction between isotropization and fine-tuning PLMs.

\subsection{Measuring isotropy}
Following \citet{DBLP:conf/iclr/MuV18, bis2021too}, we adopt the following measurement of isotropy:
\begin{equation}
    \begin{split}
        \text{I}(\mathbf{V}) = \frac{\min_{\mathbf{c}~\in~C}\text{Z}(\mathbf{c}, \mathbf{V})}{\max_{\mathbf{c}~\in~C}\text{Z}(\mathbf{c}, \mathbf{V})},
    \end{split}
    \label{equation: transfer learning task}
\end{equation}
where $\mathbf{V} \in \mathbb{R}^{N \times d} $ is the matrix of stacked embeddings of $N$ utterances (note that the embeddings have zero mean), $C$ is the set of unit eigenvectors of $\mathbf{V}^\top \mathbf{V}$, and $\text{Z}(\mathbf{c}, \mathbf{V})$ is the partition function~\cite{arora-etal-2016-latent} defined as:
\begin{equation}
    \begin{split}
        \text{Z}(\mathbf{c}, \mathbf{V}) = \sum_{i=1}^{N}{\exp{\left({\mathbf{c}^\top\mathbf{v}_i}\right)}},
    \end{split}
    \label{equation: transfer learning task}
\end{equation}
where $\mathbf{v}_i$ is the $i_{\text{th}}$ row of $\mathbf{V}$.
$\text{I}(\mathbf{V})\in\left[0,1\right]$, and 1 indicates perfect isotropy.


\begin{table}[t]
\centering
\small
\begin{tabular}{lccc}
\toprule
 Dataset & BERT & IntentBERT \\
\midrule
BANKING  &  .96 & .71\tiny{(.04)} \\
HINT3    &  .95 & .72\tiny{(.03)} \\
HWU64    &  .96 & .72\tiny{(.04)} \\
\bottomrule
\end{tabular}
\caption{The impact of fine-tuning on isotropy. Fine-tuning renders the semantic space notably more anisotropic. The mean and standard deviation of 5 runs with different random seeds are reported.
}
\label{table: pilot, fine-tune affects isotropy.}
\end{table}

\subsection{Fine-tuning Leads to Anisotropy}
\label{section: pilot experiments, fine-tuning -> anisotropy}
To observe the impact of fine-tuning on isotropy, we follow IntentBERT~\cite{zhang-etal-2021-effectiveness-pre} to fine-tune BERT~\cite{devlin2018bert} with standard supervised training on a small set of an intent detection benchmark OOS~\cite{larson2019evaluation} (details are given in Section~\ref{section_methodoloy_supervised_pre-training}). We then compare the isotropy of the original embedding space (BERT) and the embedding space after fine-tuning (IntentBERT) on target datasets. As shown in Table~\ref{table: pilot, fine-tune affects isotropy.}, after fine-tuning, the isotropy of the embedding space is notably decreased on all datasets.
Hence, it can be seen that \emph{fine-tuning may render the feature space more anisotropic}. 


\begin{figure}[!h]
    \centering
    \includegraphics[scale=0.45]{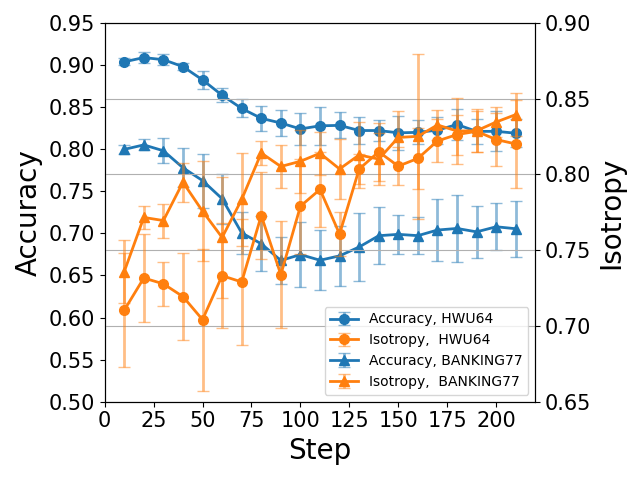}
    \caption{The impact of contrastive learning on IntentBERT with experiments on HWU64 and BANKING77 datasets. The performance (blue) drops while the isotropy (orange) increases.}
    \label{figure: twoStage_contrastiveLearning_hwu64}
\end{figure}
\subsection{Isotropization after Fine-tuning May Have an Adverse Effect }
\begin{figure}[h]
\centering
     \begin{subfigure}[b]{0.45\textwidth}
         \centering
         \includegraphics[scale = 0.4]{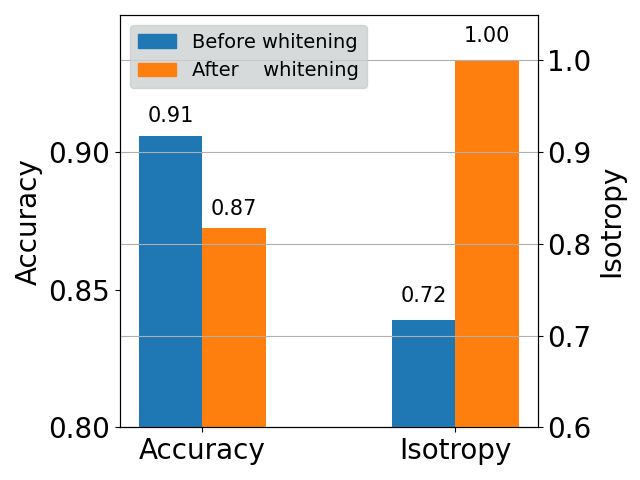}
         \caption{HWU64.}
         \label{figure: twoStage_whitening_hwu64_banking7: HWU64}
     \end{subfigure}
     \\
     \begin{subfigure}[b]{0.45\textwidth}
         \centering
         \includegraphics[scale = 0.4]{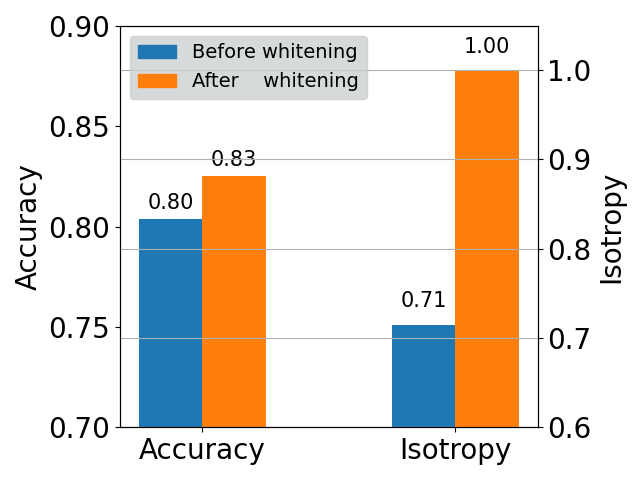}
         \caption{BANKING77.}
         \label{figure: twoStage_whitening_hwu64_banking7: bank77}
     \end{subfigure}
\caption{The impact of whitening on IntentBERT with experiments on HWU64 and BANKING77 datasets. Whitening transformation leads to perfect isotropy but has inconsistent effects on the performance.}
\label{figure: twoStage_whitening_hwu64_banking77}
\end{figure}

To examine the effect of isotropization on a fine-tuned model, we apply two strong isotropization techniques to IntentBERT: dropout-based contrastive learning~\cite{gao2021simcse} and whitening transformation~\cite{su2021whitening}. The former fine-tunes PLMs in a contrastive learning manner\footnote{We refer the reader to the original paper for details.}, while the latter transforms the semantic feature space into an isotropic space via matrix transformation. 
These methods have been demonstrated highly effective~\cite{gao2021simcse, su2021whitening}
when applied to off-the-shelf PLMs, but things are different when they are applied to fine-tuned models. 
As shown in Fig.~\ref{figure: twoStage_contrastiveLearning_hwu64}, contrastive learning improves isotropy, but it significantly lowers the performance on two benchmarks.
As for whitening transformation, it has inconsistent effects on the two datasets,  
as shown in Fig.~\ref{figure: twoStage_whitening_hwu64_banking77}. It hurts the performance on HWU64~(Fig.~\ref{figure: twoStage_whitening_hwu64_banking7: HWU64}) but yields better results on BANKING77~(Fig.~\ref{figure: twoStage_whitening_hwu64_banking7: bank77}), while producing nearly perfect isotropy on both.
The above observations indicate that \emph{isotropization may hurt fine-tuned models}, which echoes the recent finding of~\citeauthor{rajaee2021does}.

\section{Method}

\begin{figure*}[h]
\centering
     \begin{subfigure}[t]{0.3\textwidth}
         \centering
         \includegraphics[scale = 0.6]{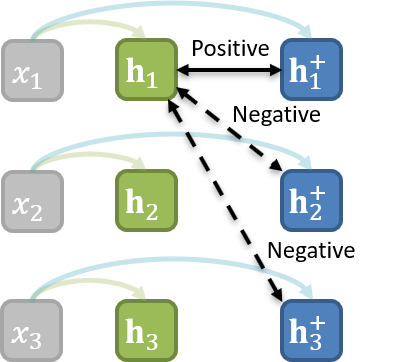}
         \caption{CL-Reg.}
         \label{subFigure: figure: method_2reg_CLReg}
     \end{subfigure}
     \begin{subfigure}[t]{0.65\textwidth}
         \centering
         \includegraphics[scale = 0.6]{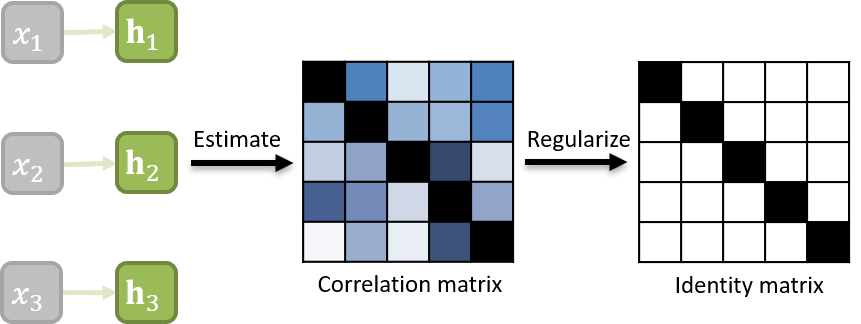}
         \caption{Cor-Reg.}
         \label{subFigure: figure: method_2reg_corReg}
     \end{subfigure}
\caption{Illustration of CL-Reg (contrastive-learning-based regularizer) and Cor-Reg (correlation-matrix-based regularizer). $x_i$ is the $i_\text{th}$ utterance in a batch of size $3$. In (a), $x_i$ is fed to the PLM twice with built-in dropout to produce two different representations of $x_i$: $\mathbf{h}_i$ and $\mathbf{h}_i^{+}$. Positive and negative pairs are then constructed for each $x_i$. For example, $\mathbf{h}_1$ and $\mathbf{h}_1^{+}$ form a positive pair for $x_1$, while $\mathbf{h}_1$ 
and $\mathbf{h}_2^{+}$, and $\mathbf{h}_1$ and $\mathbf{h}_3^{+}$, form negative pairs for $x_1$. In (b), the correlation matrix is estimated from $\textbf{h}_i$, feature vectors generated by the PLM, and is regularized towards the identity matrix.}
\label{figure: method_2reg}
\end{figure*}
\label{section_methodoloy}
The pilot experiments reveal the anisotropy of a PLM fine-tuned on intent detection tasks and the challenge of applying isotropization techiniques on the fine-tuned model. In this section, we propose a joint fine-tuning and isotropization framework. Specifically, we propose two regularizers to make the feature space more isotropic during fine-tuning. Before presenting our method, we first introduce supervised pre-training.

\subsection{Supervised Pre-training for Few-shot Intent Detection}
\label{section_methodoloy_supervised_pre-training}
Few-shot intent detection targets to train a good intent classifier with only a few labeled data $\mathcal{D}_{\text{target}}=\{(x_i,y_i)\}_{N_t}$, where $N_t$ is the number of labeled samples in the target dataset, $x_i$ denotes the $i_\text{th}$ utterance, and $y_i$ is the label.

To tackle the problem, \citet{zhang-etal-2021-effectiveness-pre} propose to learn intent detection skills (fine-tune a PLM) on a small subset of public intent detection benchmarks by supervised pre-training. Denote by $\mathcal{D}_{\text{source}}=\{(x_i,y_i)\}_{N_s}$ the source data used for pre-training, where $N_s$ is the number of examples. The fine-tuned PLM can be directly used on the target dataset. It has been shown that this method can work well  when 
the label spaces of $\mathcal{D}_{\text{source}}$ and $\mathcal{D}_{\text{target}}$ are disjoint. 

Specifically, the pre-training is conducted by attaching a linear layer (as the classifier) on top of the utterance representation generated by the PLM: 
\begin{equation}
    \begin{split}
        \text{p}(y|\mathbf{h}_i) = \text{softmax}\left(\mathbf{W}\mathbf{h}_i + \mathbf{b}\right) \in \mathbb{R}^L,
    \end{split}
    \label{equation: linear layer classifier}
\end{equation}
where $\mathbf{h}_i \in \mathbb{R}^{d}$ is the representation of the $i_\text{th}$ utterance in $\mathcal{D}_{\text{source}}$, $\mathbf{W} \in \mathbb{R}^{L \times d}$ and $\mathbf{b} \in \mathbb{R}^{L}$ are the parameters of the linear layer, and $L$ is the number of classes. The model parameters $\theta=\left\{\phi, \mathbf{W}, \mathbf{b}\right\}$, with $\phi$ being the parameters of the PLM, are trained on $\mathcal{D}_{\text{source}}$ with a cross-entropy loss:
\begin{equation}
        \theta = \argmin_{\theta} \mathcal{L}_{\text{ce}}\left(\mathcal{D}_{\text{source}};\theta\right).
    \label{equation: transfer learning task}
\end{equation}
After supervised pre-training, the linear layer is removed, and the PLM can be immediately used as a feature extractor for few-shot intent classification on target data. As shown in \citet{zhang-etal-2021-effectiveness-pre}, a parametric classifier such as logistic regression can be trained with only a few labeled samples to achieve good performance. 

However, our analysis in Section~\ref{section: pilot experiments, fine-tuning -> anisotropy} shows the limitation of supervised pre-training, which yields a anisotropic feature space.

\subsection{Regularizing Supervised Pre-training with Isotropization}
To mitigate the anisotropy of the PLM fine-tuned by supervised pre-training, we propose a joint training objective by adding a regularization term  $\mathcal{L}_{\text{reg}}$ for isotropization:
\begin{equation}
    \mathcal{L}=\mathcal{L}_{\text{ce}}(\mathcal{D}_{\text{source}};\theta)+\lambda \mathcal{L}_{\text{reg}}(\mathcal{D}_{\text{source}};\theta),
    \label{equation: loss, general isotropy regularizer}
\end{equation}
where $\lambda$ is a weight parameter. The aim is to learn intent detection skills while maintaining an appropriate degree of isotropy. We devise two different regularizers introduced as follows.

\textbf{Contrastive-learning-based Regularizer.} Inspired by the recent success of contrastive learning in mitigating anisotropy~\cite{yan-etal-2021-consert, gao2021simcse}, we employ the dropout-based contrastive learning loss used in \citet{gao2021simcse} as the regularizer:
\begin{equation}
    \mathcal{L}_{\text{reg}}=-\frac{1}{N_b}\sum_i^{N_b}\text{log}\frac{e^{\text{sim}(\mathbf{h}_i, \mathbf{h}_{i}^{+})/\tau}}{\sum_{j=1}^{N_b}e^{\text{sim}(\mathbf{h}_i, \mathbf{h}_{j}^{+})/\tau}}.
    \label{equation: regularizer contrastive}
\end{equation}
In particular, $\mathbf{h}_i \in \mathbb{R}^{d}$ and $\mathbf{h}_i^+ \in \mathbb{R}^{d}$ are two different representations of utterance $x_i$ generated by the PLM with built-in standard dropout~\cite{JMLR:v15:srivastava14a}, i.e., $x_i$ is passed to the PLM twice with different dropout masks to produce $\mathbf{h}_i$ and $\mathbf{h}_i^+$. $\text{sim}(\mathbf{h}_1, \mathbf{h}_2)$ denotes the cosine similarity between $\mathbf{h}_1$ and $\mathbf{h}_2$. $\tau$ is the temperature parameter. $N_b$ is the batch size.  
Since $\mathbf{h}_i$ and $\mathbf{h}_i^+$ represent the same utterance, they form a positive pair. Similarly, $\mathbf{h}_i$ and $\mathbf{h}_j^+$ form a negative pair, since they represent  different utterances. An example is given in Fig.~\ref{subFigure: figure: method_2reg_CLReg}. By minimizing the contrastive loss, positive pairs are pulled together while negative pairs are pushed away, which in theory enforces an isotropic feature space~\cite{gao2021simcse}.  
In \citet{gao2021simcse}, the contrastive loss is used as the single objective to fine-tune off-the-shelf PLMs in an unsupervised manner, while in this work we use it jointly with supervised pre-training to fine-tune PLMs for few-shot learning.

\label{section-method-cor-reg}
\textbf{Correlation-matrix-based Regularizer.} The above regularizer enforces isotropization implicitly. Here, we propose a new regularizer that explicitly enforces isotropization.
The perfect isotropy is characterized by zero covariance and uniform variance~\cite{su2021whitening, Zhou2021IsoBNFB}, i.e., a covariance matrix with uniform diagonal elements and zero non-diagonal elements. Isotropization can be achieved by endowing the feature space with such statistical property. However, as will be shown in Section~\ref{section: analysis, compared to covariance}, it is difficult to determine the  appropriate scale of variance. Therefore, we base the regularizer on \emph{correlation matrix} 
:
\begin{equation}
    \mathcal{L}_{\text{reg}}=\lVert\mathbf{\Sigma} - \mathbf{I}\rVert,
    \label{equation: regularizer correlation}
\end{equation}
where $\lVert\cdot\rVert$ denotes Frobenius norm, $\mathbf{I} \in \mathbb{R}^{d \times d}$ is the identity matrix, $\mathbf{\Sigma} \in \mathbb{R}^{d \times d}$ is the correlation matrix with $\mathbf{\Sigma}_{ij}$ being the Pearson correlation coefficient between the $i_\text{th}$ dimension and the $j_\text{th}$ dimension. As shown in Fig.~\ref{subFigure: figure: method_2reg_corReg}, $\mathbf{\Sigma}$ is estimated with utterances in the current batch. By pushing the correlation matrix towards the identity matrix during training, we can learn a more isotropic feature space.




Moreover, the proposed two regularizers can be used together as follows:
\begin{equation}
    \begin{split}
        \mathcal{L}=\mathcal{L}_{\text{ce}}(\mathcal{D}_{\text{source}};\theta)+\lambda_1 \mathcal{L}_{\text{cl}}(\mathcal{D}_{\text{source}};\theta) \\ +\lambda_2 \mathcal{L}_{\text{cor}}(\mathcal{D}_{\text{source}};\theta),
    \end{split}
    \label{equation: loss, general isotropy regularizer}
\end{equation}
where $\lambda_1$ and $\lambda_2$ are the weight parameters, and $\mathcal{L}_{\text{cl}}$ and $\mathcal{L}_{\text{cor}}$ denote CL-Reg and Cor-Reg, respectively. Our experiments show that better performance is often observed when they are used together.

\section{Experiments}
To validate the effectiveness of the approach, we conduct extensive experiments.

\subsection{Experimental Setup}
\label{subsection:cross domain}
\textbf{Datasets.} To perform supervised pre-training, we follow ~\citeauthor{zhang-etal-2021-effectiveness-pre} to use the  OOS dataset~\cite{larson2019evaluation} which contains diverse semantics of $10$ domains. 
Also following ~\citeauthor{zhang-etal-2021-effectiveness-pre}, we exclude the domains ``Banking'' and ``Credit Cards'' 
since they are similar in semantics to one of the test dataset BANKING77. We then use $6$ domains for training and $2$ for validation, as shown in Table~\ref{table: domain split of OOS}. For evaluation, we employ three datasets:
\textbf{BANKING77}~\cite{casanueva2020efficient} is an intent detection dataset for banking service.
\textbf{HINT3}~\cite{arora2020hint3} covers $3$ domains, ``Mattress Products Retail'', ``Fitness Supplements Retail'', and ``Online Gaming''. \textbf{HWU64}~\cite{DBLP:conf/iwsds/LiuESR19} is a large-scale dataset containing $21$ domains. Dataset statistics are summarized in Table~\ref{table: Dataset statistics}.
\begin{table}[h]
\centering
\small
\begin{tabular}{p{0.2\textwidth}p{0.2\textwidth}}
\toprule
Training &  Validation
\\
\midrule
``Utility'', ``Auto commute'', ``Work'', ``Home'', ``Meta'', ``Small talk'' & ``Travel'', ``Kitchen dining'' \\
\bottomrule
\end{tabular}
\caption{Split of domains in OOS.}
\label{table: domain split of OOS}
\end{table}


\begin{table}[h]
\centering
\small
\begin{tabular}{lccc}
\toprule
Dataset   &  \#domain & \#intent & \#data \\
\midrule
OOS       &   10   & 150   & 22500    \\
BANKING77 &   1    & 77    & 13083    \\
HINT3     &   3    & 51    & 2011     \\  
HWU64     &   21   & 64    & 10030    \\
\bottomrule
\end{tabular}
\caption{Dataset statistics.}
\label{table: Dataset statistics}
\end{table}

\textbf{Our Method.}
Our method 
can be applied to fine-tune any PLM. We conduct experiments on two popular PLMs, BERT~\cite{devlin2018bert} and RoBERTa~\cite{liu2019roberta}. For both of them, the embedding of $[CLS]$ is used as the utterance representation in Eq.~\ref{equation: linear layer classifier}. We employ logistic regression as the classifier. We select the hyperparameters $\lambda, \lambda_1, \lambda_2$, and $\tau$ by validation. The best hyperparameters are provided in Table~\ref{table: best hyperparameters}.

\begin{table}[h]
\begin{subtable}{0.48\textwidth}
\small
\centering
\begin{tabular}{lc}
\toprule
Method & Hyperparameter \\
\midrule
CL-Reg           & $\lambda=1.7, \tau=0.05$ \\
Cor-Reg          & $\lambda=0.04$ \\
CL-Reg + Cor-Reg & $\lambda_1=1.7, \lambda_2=0.04, \tau=0.05$ \\
\bottomrule
\end{tabular}
\caption{BERT-based.}
\end{subtable}\\
\newline
\newline
\begin{subtable}{0.48\textwidth}
\small
\centering
\begin{tabular}{lc}
\toprule
Method & Hyperparameter \\
\midrule
CL-Reg           & $\lambda=2.9, \tau=0.05$ \\
Cor-Reg          & $\lambda=0.06$ \\
CL-Reg + Cor-Reg & $\lambda_1=2.9, \lambda_2=0.13, \tau=0.05$ \\
\bottomrule
\end{tabular}
\caption{RoBERTa-based.}
\end{subtable}

\caption{Hyperparameters selected via validation.}
\label{table: best hyperparameters}
\end{table}



\begin{table*}[t]
\centering
\small
\begin{tabular}{lcccccccc}
\toprule
\multicolumn{1}{l}{\multirow{2}{*}{Method}} & 
\multicolumn{2}{c}{BANKING77} &
\multicolumn{2}{c}{HINT3} &
\multicolumn{2}{c}{HWU64} &
\multicolumn{2}{c}{Val.}  
\\
\cmidrule(lr){2-3}  \cmidrule(lr){4-5}  \cmidrule(lr){6-7} \cmidrule(lr){8-9}
& 2-shot & 10-shot & 2-shot & 10-shot & 2-shot & 10-shot & 2-shot & 10-shot\\
\midrule
BERT-Freeze & 57.10 & 84.30 & 51.95 & 80.27 & 64.83  & 87.99   & 74.20 & 92.99  \\
CONVBERT$\ssymbol{5}$ & 68.30 & 86.60 & 72.60 & 87.20 & 81.75 & 92.55   & 90.54 & 96.82  \\
TOD-BERT$\ssymbol{5}$ & 77.70 & 89.40 & 68.90 &  83.50 & 83.24 & 91.56   & 88.10 & 96.39   \\
USE-ConveRT$\ssymbol{5}$ & -- & 85.20 & -- & -- & --   & 85.90 & -- & --  \\
DNNC-BERT$\ssymbol{5}$ & 67.50 & 89.80 & 64.10 & 87.90  & 73.97 & 90.71   & 72.98 & 95.23  \\
CPFT-BERT & 72.09 & 89.82 & 74.34 & 90.37 & 83.02 & 93.66 & 89.33 & 97.30 \\
IntentBERT$\ssymbol{5}$ & 82.40 & 91.80 & \textbf{80.10} & 90.20 & -- & -- & -- & -- \\
IntentBERT-ReImp & 80.38\tiny{(.35)} & 92.35\tiny{(.12)} & 77.09\tiny{(.89)} & 89.55\tiny{(.63)} & 90.61\tiny{(.44)}  & 95.21\tiny{(.15)}  & 93.62\tiny{(.38)} & 97.80\tiny{(.18)} \\
BERT-White & 72.95  & 88.86 & 65.70 & 85.70  & 75.98 & 91.26 & 87.33  & 96.05 \\
IntentBERT-White & 82.52\tiny{(.26)} & 92.29\tiny{(.33)} & 78.50\tiny{(.59)} & 90.14\tiny{(.26)} & 87.24\tiny{(.18)}  & 94.42\tiny{(.08)}  & \textbf{94.89\tiny{(.21)}} & 98.07\tiny{(.12)} \\
\midrule
CL-Reg & \textbf{83.45\tiny{(.35)}} & \textbf{93.66\tiny{(.22)}} & 79.30\tiny{(.87} & \textbf{91.06\tiny{(.30)}} &  \textbf{91.46\tiny{(.15)}} &  \textbf{95.84\tiny{(.12)}} & 94.43\tiny{(.22)} & \textbf{98.43\tiny{.02)}} \\
Cor-Reg  & \textbf{83.94\tiny{(.45)}} & \textbf{93.98\tiny{(.26)}} & \textbf{80.16\tiny{(.71)}} & \textbf{91.38\tiny{(.55)}} & \textbf{90.75\tiny{(.35)}} & \textbf{95.82\tiny{(.14)}} & \textbf{95.02\tiny{(.22)}} & \textbf{98.47\tiny{(.07)}} \\
CL-Reg + Cor-Reg   & \textbf{85.21\tiny{(.58)}} & \textbf{94.68\tiny{(.01)}} & \textbf{81.20\tiny{(.45)}} & \textbf{92.38\tiny{(.01)}} & \textbf{90.66\tiny{(.42)}} & \textbf{95.84\tiny{(.19)}} & \textbf{95.41\tiny{(.25)}} & \textbf{98.58\tiny{(.01)}} \\
\bottomrule
\end{tabular}
\caption{$5$-way few-shot intent detection using BERT. We report the mean and standard deviation of our methods and IntentBERT variants. CL-Reg, Cor-Reg, and CL-Reg + CorReg denote supervised pre-training regularized by the corresponding regularizer. The top $3$ results are highlighted. $\ssymbol{5}$ denotes results from \cite{zhang-etal-2021-effectiveness-pre}.}
\label{table: main_result_effectiveness_of_our_methods_BERT}
\end{table*}
\begin{table*}[t]
\centering
\small
\begin{tabular}{lcccccccc}
\toprule
\multicolumn{1}{l}{\multirow{2}{*}{Method}} & 
\multicolumn{2}{c}{BANKING77} &
\multicolumn{2}{c}{HINT3} &
\multicolumn{2}{c}{HWU64} &
\multicolumn{2}{c}{Val.}  
\\
\cmidrule(lr){2-3}  \cmidrule(lr){4-5}  \cmidrule(lr){6-7} \cmidrule(lr){8-9}
& 2-shot & 10-shot & 2-shot & 10-shot & 2-shot & 10-shot & 2-shot & 10-shot\\
\midrule
RoBERTa-Freeze & 60.74 & 82.18 & 57.90 & 79.26 & 75.30 & 89.71   & 74.86 & 90.52 \\
WikiHowRoBERTa & 32.88 & 59.50 & 31.92 & 54.18 & 30.81 & 52.47   & 34.10 & 60.59 \\
DNNC-RoBERTa  &   74.32 & 87.30 & 68.06 & 82.34 & 69.87 & 80.22 & 58.51 & 74.46 \\
CPFT-RoBERTa & 80.27\tiny{(.11)} & 93.91\tiny{(.06)} & 79.98\tiny{(.11)} & \textbf{92.55\tiny{(.07)}} & 83.18\tiny{(.11)} & 92.82\tiny{(.06)} & 86.71\tiny{(.10)} & 96.45\tiny{(.05)} \\
IntentRoBERTa & 81.38\tiny{(.66)} & 92.68\tiny{(.24)} & 78.20\tiny{(1.72)} & 89.01\tiny{(1.07)} & \textbf{90.48\tiny{(.69)}} & 94.49\tiny{(.43)} & 95.33\tiny{(.54)} & 98.32\tiny{(.15)} \\
RoBERTa-White & 79.27  & 93.00  & 73.13  & 89.02  & 82.65  & 94.00 & 89.90  & 97.14  \\
IntentRoBERTa-White & 83.75\tiny{(.45)} & 92.68\tiny{(.31)} & 79.64\tiny{(1.38)} & 90.13\tiny{(.66)} & 86.52\tiny{(1.33)} & 93.82\tiny{(.53)} & 96.06\tiny{(.58)} & 98.35\tiny{(.21)} \\
\midrule
CL-Reg  & \textbf{84.63\tiny{(.68)}} & \textbf{94.43\tiny{(.34)}} & \textbf{81.10\tiny{(.49)}} & 91.65\tiny{(.13)} & \textbf{91.67\tiny{(.20)}} & \textbf{95.44\tiny{(.28)}} & \textbf{96.32\tiny{(.14)}} & \textbf{98.79\tiny{(.05)}} \\
Cor-Reg & \textbf{86.92\tiny{(.71)}} & \textbf{95.07\tiny{(.41)}} & \textbf{82.20\tiny{(.48)}} & \textbf{92.11\tiny{(.41)}} & \textbf{91.10\tiny{(.18)}} & \textbf{95.69\tiny{(.12)}} & \textbf{96.82\tiny{(.03)}} & \textbf{98.89\tiny{(.03)}} \\
CL-Reg + Cor-Reg & \textbf{87.96\tiny{(.31)}} & \textbf{95.85\tiny{(.02)}} & \textbf{83.55\tiny{(.30)}} & \textbf{93.17\tiny{(.23)}} & 90.47\tiny{(.39)} & \textbf{95.64\tiny{(.28)}} & \textbf{96.35\tiny{(.19)}} & \textbf{98.85\tiny{(.07)}} \\
\bottomrule
\end{tabular}
\caption{$5$-way few-shot intent detection using RoBERTa. We report the mean and standard deviation of our methods and IntentBERT variants. CL-Reg, Cor-Reg, and CL-Reg + CorReg denote supervised pre-training regularized by the corresponding regularizer. The top $3$ results are highlighted.}
\label{table: main_result_effectiveness_of_our_methods_RoBERTa}
\end{table*}

\textbf{Baselines.} We compare our method to the following baselines. First, for BERT-based methods, \textbf{BERT-Freeze} freezes BERT; \textbf{CONVBERT}~\cite{mehri2020dialoglue}, \textbf{TOD-BERT}~\cite{wu2020tod}, and \textbf{DNNC-BERT}~\cite{zhang-etal-2020-discriminative} further pre-train BERT on conversational corpus or natural language inference tasks.
\textbf{USE-ConveRT}~\cite{henderson-etal-2020-convert, casanueva2020efficient} is a transformer-based dual-encoder pre-trained on conversational corpus.
\textbf{CPFT-BERT} is the re-implemented version of CPFT~\cite{zhang-etal-2021-shot}, by further pre-training BERT in an unsupervised manner with mask-based contrastive learning and masked language modeling on the same training data as ours. 
\textbf{IntentBERT}~\cite{zhang-etal-2021-effectiveness-pre} further pre-trains BERT via supervised pre-training described in Section~\ref{section_methodoloy_supervised_pre-training}. 
To guarantee a fair comparison, we provide \textbf{IntentBERT-ReImp}, the re-implemented version of IntentBERT, which uses the same random seed, training data, and validation data as our methods. Second, for RoBERTa-based baselines, \textbf{RoBERTa-Freeze} freezes the model. \textbf{WikiHowRoBERTa}~\cite{zhang2020intent} further pre-trains RoBERTa on synthesized intent detection data.
\textbf{DNNC-RoBERTa} and \textbf{CPFT-RoBERTa} are similar to DNNC-BERT and CPFT-BERT except the PLM.
\textbf{IntentRoBERTa} is the re-implemented version of IntentBERT based on RoBERTa, with uses the same random seed, training data, and validation data as our method.
Finally, to show the superiority of the joint fine-tuning and isotropization, we compare our method against whitening transformation~\cite{su2021whitening}. \textbf{BERT-White} and \textbf{RoBERTa-White} apply the transformation to BERT and RoBERTa, respectively. \textbf{IntentBERT-White} and \textbf{IntentRoBERTa-White} apply the transformation to IntentBERT-ReImp and IntentRoBERTa, respectively.

All baselines use logistic regression as classifier except DNNC-BERT and DNNC-RoBERTa, wherein we follow the original work\footnote{https://github.com/salesforce/DNNC-few-shot-intent} to 
train a pairwise encoder for nearest neighbor classification.

\textbf{Training Details.} We use PyTorch library and Python 
to build our model.
We employ Hugging Face implementation\footnote{https://github.com/huggingface/transformers} of \emph{bert-base-uncased} and \emph{roberta-base}. We use Adam~\cite{kingma2014adam} as the optimizer with learning rate of $2e-05$ and weight decay of $1e-03$. The model is trained with Nvidia RTX 3090 GPUs. The training is early stopped if no improvement in validation accuracy is observed for $100$ steps. The same set of random seeds, $\left\{1, 2, 3, 4, 5\right\}$, is used for IntentBERT-ReImp, IntentRoBERTa, and our method.




\textbf{Evaluation.} 
The baselines and our method are evaluated on $C$-way $K$-shot tasks. For each task, we randomly sample $C$ classes and $K$ examples per class. The $C\times K$ labeled examples are used to train the logistic regression classifier. Note that we do not further fine-tune the PLM using the labeled data of the task.
We then sample another $5$ examples per class as queries. Fig.~\ref{figure: idea_illustration} gives an example with $C=2$ and $K=1$. We report the averaged accuracy of $500$ tasks randomly sampled from $\mathcal{D}_{\text{target}}$.

\subsection{Main Results}
\label{main_results}
The main results are provided in Table~\ref{table: main_result_effectiveness_of_our_methods_BERT}~(BERT-based) and Table~\ref{table: main_result_effectiveness_of_our_methods_RoBERTa}~(RoBERTa-based). The following observations can be made. First, 
our proposed regularized supervised pre-training, with either CL-Reg or Cor-Reg, consistently outperforms all the baselines by a notable margin in most cases, indicating the effectiveness of our method.
Our method also outperforms whitening transformation, demonstrating the superiority of the proposed joint fine-tuning and isotropization framework. Second, Cor-Reg slightly outperforms CL-Reg in most cases, showing the advantage of 
enforcing isotropy explicitly with the correlation matrix. Finally, CL-Reg and Cor-Reg show a complementary effect in many cases, especially on BANKING77. The above observations are consistent for both BERT and RoBERTa. It can be also seen that higher performance is often attained with RoBERTa.

\begin{table}[h]
\renewrobustcmd{\bfseries}{\fontseries{b}\selectfont}
\renewrobustcmd{\boldmath}{}
\centering
\small
\begin{tabular}{wl{2.5cm}wc{1.5cm}wc{1.0cm}wc{1.0cm}}
\toprule
Method & BANKING77 & HINT3 & HWU64 \\
\midrule
IntentBERT-ReImp   & .71\tiny{(.04)}  &  .72\tiny{(.03)}  & .72\tiny{(.03)}  \\
SPT+CL-Reg   & .77\tiny{(.01)}   &  .78\tiny{(.01)}  &   .75\tiny{(.03)}  \\
SPT+Cor-Reg  & 
 .79\tiny{(.01)}  & .76\tiny{(.06)}   &     .80\tiny{(.03)}  \\
SPT+CL-Reg+Cor-Reg  & 
 .79\tiny{(.01)}  & .76\tiny{(.05)}   &     .80\tiny{(.02)}  \\
\bottomrule
\end{tabular}
\caption{Impact of the proposed regularizers on isotropy. The results are obtained with BERT. SPT denotes supervised pre-training.}
\label{table_analysis_isotropy_change_English}
\end{table}

\label{section: analysis, better isotropy}
The observed improvement in performance comes with an improvement in isotropy.
We report the change in isotropy by the proposed regularizers in Table~\ref{table_analysis_isotropy_change_English}. It can be seen that both regularizers and their combination make the feature space more isotropic compared to  IntentBERT-ReImp that  only uses supervised pre-training. In addition, in general, Cor-Reg can achieve better isotropy than CL-Reg.


\subsection{Ablation Study and Analysis}
\label{section: analysis}
\begin{figure}[ht]
    \centering
    \includegraphics[scale=0.45]{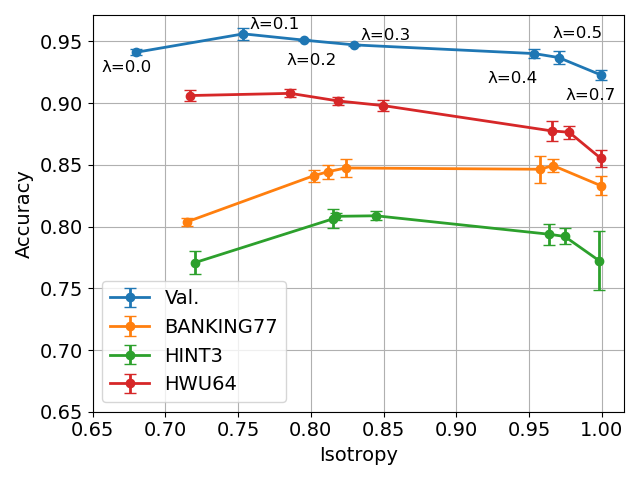}
    \caption{Relation between performance and isotropy. The results are obtained with BERT on $5$-way $2$-shot tasks.}
    \label{figure: acc_iso_curve}
\end{figure}

\textbf{Moderate isotropy is helpful.}
To investigate the relation between the isotropy of the feature space and the performance of few-shot intent detection, we tune the weight parameter $\lambda$ of Cor-Reg to 
increase the isotropy and examine the performance. As shown in Fig.~\ref{figure: acc_iso_curve}, a common pattern is observed: the best performance is achieved when the isotropy is moderate. 
This observation indicates that it is important to find an appropriate trade-off between learning intent detection skills and learning an insotropic feature space. In our method, we select the appropriate $\lambda$ by validation.

\textbf{Correlation matrix is better than covariance matrix as regularizer.}
\label{section: analysis, compared to covariance}
In the design of Cor-Reg (Section~\ref{section-method-cor-reg}), we use the correlation matrix, rather than the covariance matrix, to characterize isotropy, although the latter contains more information -- variance. The reason is that it is difficult to determine the proper scale of the variances. 
Here, we conduct experiments using the covariance matrix, by pushing the non-diagonal elements (covariances) towards $0$ and the diagonal elements (variances) towards $1$, $0.5$, or the mean value, which are denoted by Cov-Reg-1, Cov-Reg-0.5, and Cov-Reg-mean respectively in Table~\ref{table: analysis, coefficient matrix, or covariance matrix}. It can be seen that all the variants perform worse than Cor-Reg.
\begin{table}[h]
\centering
\small
\begin{tabular}{lcc}
\toprule
Method & BANKING77 & Val. \\
\midrule
Cov-Reg-1    & 82.19\tiny{(.84)} & 94.52\tiny{(.19)} \\
Cov-Reg-0.5  & 82.62\tiny{(.80)} & 94.52\tiny{(.26)} \\
Cov-Reg-mean & 82.50\tiny{(1.00)} & 93.82\tiny{(.39)} \\
\midrule
Cor-Reg (ours) & \textbf{83.94\tiny{(.45)}} & \textbf{95.02\tiny{(.22)}} \\
\bottomrule
\end{tabular}
\caption{
Comparison between using covariance matrix and using  correlation matrix to implement Cor-Reg. The experiments are conducted with BERT and evaluated on $5$-way $2$-shot tasks.}
\label{table: analysis, coefficient matrix, or covariance matrix}
\end{table}


\textbf{Our method is complementary with batch normalization.} Batch normalization~\cite{ICML-2015-IoffeS} can potentially mitigate the anisotropy problem via normalizing each dimension 
with unit variance.
We find that combining our method with batch normalization yields better performance, as shown in Table~\ref{table: analysis, batch normalization}.
\begin{table}[h]
\centering
\small
\begin{tabular}{ccccc}
\toprule
SPT & CL-Reg & Cor-Reg & BN & BANKING77 \\
\midrule
\checkmark &  &  &  & 80.38\tiny(.35) \\
\checkmark &  &  & \checkmark  & 82.38\tiny{(.38)} \\
\midrule
\checkmark & \checkmark &  &  & 83.45\tiny{(.35)} \\
\checkmark & \checkmark &  & \checkmark  & \textbf{84.18\tiny{(.28)}} \\
\midrule
\checkmark &  &  \checkmark &  & 83.94\tiny{(.45)} \\
\checkmark & &  \checkmark  & \checkmark  & \textbf{84.67\tiny{(.51)}} \\
\midrule
\checkmark & \checkmark  &  \checkmark &  & 85.21\tiny{(.58)} \\
\checkmark & \checkmark  &  \checkmark  & \checkmark  & \textbf{85.64\tiny{(.41)}} \\
\bottomrule
\end{tabular}
\caption{
Effect of combining batch normalization and our method. The experiments are conducted with BERT and evaluated on $5$-way $2$-shot tasks. SPT denotes supervised pre-training. BN denotes batch normalization.}
\label{table: analysis, batch normalization}
\end{table}

\begin{figure}[ht]
    \centering
    \includegraphics[scale=0.45]{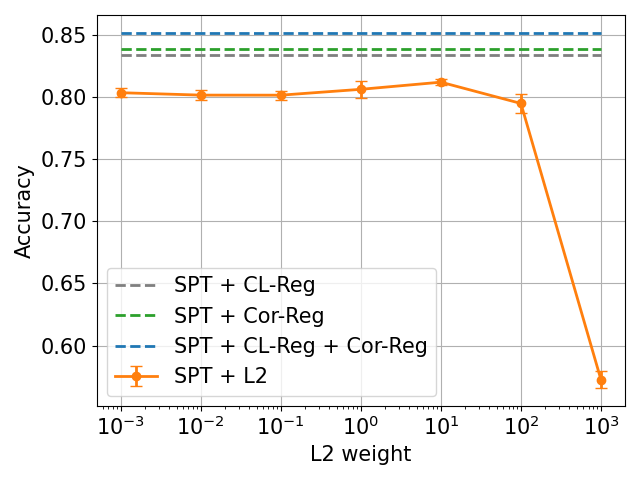}
    \caption{Comparison between our methods and L2 regularization. 
    The experiments are conducted with BERT and evaluated on $5$-way $2$-shot tasks on BANKING77. SPT denotes superivsed pre-training.}
    \label{figure: analysis_l2reg}
\end{figure}

\textbf{The performance gain is not from the reduction in model variance.} Regularization techniques such as L1 regularization~\cite{tibshirani1996regression} and L2 regularization~\cite{hoerl1970ridge} are often used to improve model performance by reducing model variance. Here, we show that the performance gain of our method is ascribed to the improved isotropy~(Table~\ref{table_analysis_isotropy_change_English}) rather than the reduction in model variance. To this end, we compare our method against L2 regularization with a wide range of weights, and it is observed that reducing model variance cannot achieve comparable performance to our method, as shown in Fig.~\ref{figure: analysis_l2reg}.

\begin{figure}[ht]
    \centering
    \includegraphics[scale=0.45]{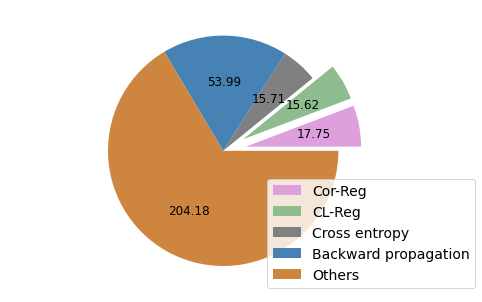}
    \caption{Run time decomposition of a single epoch. The unit is second.}
    \label{figure:computational overhead}
\end{figure}
\textbf{The computational overhead is small.} To analyze the computational overheads incurred by CL-Reg and Cor-Reg, we decompose the duration of one epoch of our method using the two regularizers jointly. As shown in Fig.~\ref{figure:computational overhead}, the overheads of CL-Reg and Cor-Reg are small,
only taking up a small portion of the time.


\section{Conclusion}
\label{sec:conclusion}
In this work, we have identified and analyzed the anisotropy of the feature space of a PLM fine-tuned on intent detection tasks. Further, we have proposed a joint training framework and designed two regularizers based on contrastive learning and correlation matrix respectively to increase the insotropy of the feature space during fine-tuning, which leads to notably improved performance on few-shot intent detection. Our findings and solutions may have broader implications for solving other natural language understanding tasks with PLM-based models. 

\section*{Acknowledgments}
We would like to thank the anonymous reviewers for their valuable comments. This research was supported by the grants of HK ITF UIM/377 and PolyU DaSAIL project P0030935 funded by RGC.



\end{document}